\def\year{2021}\relax
\documentclass[letterpaper]{article} 
\usepackage{aaai21}  
\usepackage{algorithm,algorithmic}
\usepackage{amsmath}
\usepackage{booktabs}
\usepackage{times}  
\usepackage{helvet} 
\usepackage{courier}  
\usepackage[hyphens]{url}  
\usepackage{graphicx} 
\usepackage{natbib}  
\usepackage{caption} 
\usepackage{subfigure}
\usepackage{tablefootnote}
\usepackage{multicol}
\usepackage{multirow}
\usepackage{amsfonts}
\usepackage[switch]{lineno}

\title{Inductive Graph Neural Networks for Spatiotemporal Kriging}
\author{

Yuankai Wu,\textsuperscript{\rm 1}
Dingyi Zhuang,\textsuperscript{\rm 1}
Aurelie Labbe,\textsuperscript{\rm 2}
Lijun Sun\textsuperscript{\rm 1}\thanks{Corresponding author.}
}
\affiliations{

    \textsuperscript{\rm 1}McGill University, Montreal, Canada\\
   \textsuperscript{\rm 2}HEC Montreal, Montreal, Canada\\
    yuankai.wu@mcgill.ca, dingyi.zhuang@mail.mcgill.ca, aurelie.labbe@hec.ca, lijun.sun@mcgill.ca

}
\begin{document}

\maketitle

\begin{abstract}
Time series forecasting and spatiotemporal kriging are the two most important tasks in spatiotemporal data analysis. Recent research on graph neural networks has made substantial progress in time series forecasting, while little attention has been paid to the kriging problem---recovering signals for unsampled locations/sensors. Most existing scalable kriging methods (e.g., matrix/tensor completion) are transductive, and thus full retraining is required when we have a new sensor to interpolate. In this paper, we develop an Inductive Graph Neural Network Kriging (IGNNK) model to recover data for unsampled sensors on a network/graph structure. To generalize the effect of distance and reachability, we generate random subgraphs as samples and reconstruct the corresponding adjacency matrix for each sample. By reconstructing all signals on each sample subgraph, IGNNK can effectively learn the spatial message passing mechanism. Empirical results on several real-world spatiotemporal datasets demonstrate the effectiveness of our model. In addition, we also find that the learned model can be successfully transferred to the same type of kriging tasks on an unseen dataset. Our results show that: 1) GNN is an efficient and effective tool for spatial kriging; 2) inductive GNNs can be trained using dynamic adjacency matrices; 3) a trained model can be transferred to new graph structures and 4) IGNNK can be used to generate virtual sensors.
\end{abstract}

\section{Introduction}

With recent advances in information and communications technologies (ICT), large-scale spatiotemporal datasets are collected from various applications, such as traffic sensing and climate monitoring. Analyzing these spatiotemporal datasets has attracted considerable attention. Time series forecasting and spatiotemporal kriging are two essential tasks in spatiotemporal analysis \cite{cressie2015statistics,bahadori2014fast}. While recent research advances in deep learning have made substantial progress in time series forecasting  \cite[e.g.,][]{rangapuram2018deep,salinas2019high, sen2019think}, little attention has been paid to the spatiotemporal kriging application. The goal of spatiotemporal kriging is to perform signal interpolation for unsampled locations given the observed signals from sampled locations during the same period. The interpolation results can produce a fine-grained and high-resolution realization of spatiotemporal data, which can be used to enhance real-world applications such as travel time estimation and disaster evaluation. In addition, a better kriging model can achieve higher estimation accuracy/reliability with fewer sensors, thus reducing the operation and maintenance cost of a sensor network.

For general spatiotemporal kriging problems (e.g., in Euclidean domains), a well-developed approach is Gaussian process (GP) regression \cite{cressie2015statistics,williams2006gaussian}, which uses a flexible kernel structure to characterize spatiotemporal correlations. However, GP has two limitations: (1) the model is computationally expensive and thus it cannot deal with large-scale datasets, and (2) it is difficult to model networked systems using existing graph kernel structures. To solve large-scale kriging problems in a networked system, graph regularized matrix/tensor completion has emerged as an effective solution \cite[e.g.,][]{bahadori2014fast,zhou2012kernelized,deng2016latent,takeuchi2017autoregressive}. Combining low-rank structure and spatiotemporal regularizers, these models can simultaneously characterize both global consistency and local consistency in the data. However, matrix/tensor completion is essentially transductive: for new sensors/nodes introduced to the network, we cannot directly apply a previously trained model; instead, we have to retrain the full model for the new graph structure even with only minor changes (i.e., after introducing a new sensor). In addition, the low-rank scheme is ineffective to accommodate time-varying/dynamic graph structures. For example, some sensors may retire over time without being replaced, and some sensors may be introduced at new locations. In such cases, the network structure itself is not consistent over time, making it challenging to utilize all full information.

\begin{figure}[!ht]
\centering
\subfigure[Training process of IGNNK]{%
\label{fig:1first}%
\includegraphics[width=\columnwidth]{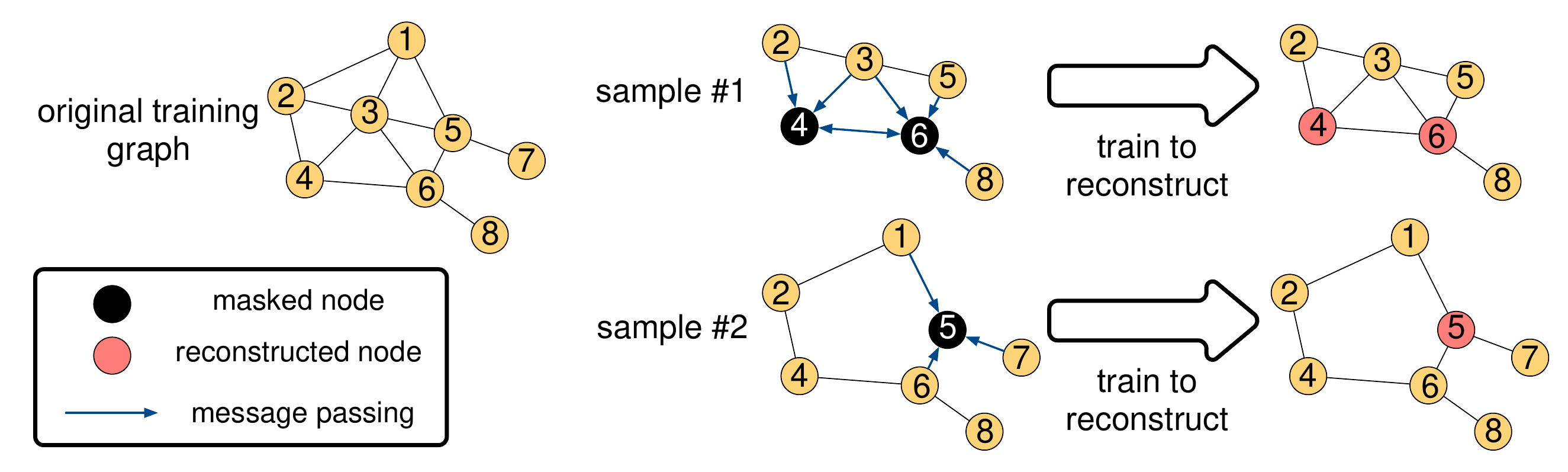}}
\subfigure[Kriging process of IGNNK]{%
\label{fig:1second}%
\includegraphics[width=\columnwidth]{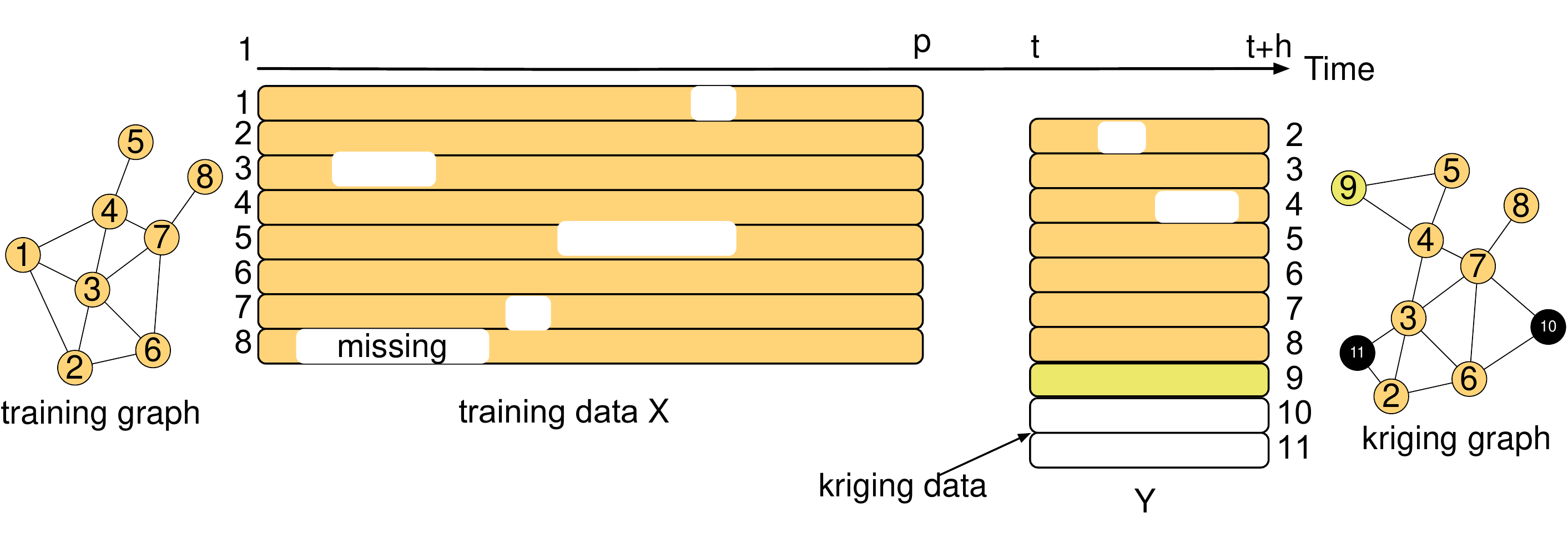}}%
\caption{Framework of of IGNNK. (a) In sample \#1, nodes $\{1,7\}$ are unsampled and nodes $\{4,6\}$ are masked and used for reconstruction. In sample \#2, The unsampled set and masked set are $\{3\}$ and $\{5\}$, respectively. (b) Illustration of real-time kriging, where the goal is to perform interpolation for virtual sensors $\{10,11\}$. Note that the set of observed sensors during $\left[t,t+h\right)$ is not necessarily the same as the set during training $\left[1,p\right]$. For example, during $\left[t,t+h\right)$, the sensor $\{1\}$ in $X$ is removed and a new sensor $\{9\}$ (in green) is added to the network. }
\label{Fig:1}
\end{figure}

Recent research has explored the potential of modeling spatiotemporal data using Graph Neural Networks (GNNs). GNNs are powerful in characterizing complex spatial dependencies by their message passing mechanism \cite{xu2018powerful}. They also demonstrate the ability and inductive power to generalize the message passing mechanism to unseen nodes or even entirely new (sub)graphs \cite{hamilton2017inductive,velivckovic2018graph,Zeng2020GraphSAINT}. Inspired by these works, here we develop an Inductive Graph Neural Network Kriging (IGNNK) model to solve real-time spatiotemporal kriging problems on dynamic network structures. Unlike graphs in recommender systems governed by certain typology, our spatial graph actually contains valuable location information which allows us to quantify the exact pairwise ``distance'' beyond ``hops''. In particular, for directed networks such as highway networks, the distance matrix will be asymmetric and actually capture the degree of ``reachability'' from one sensor to another \cite{li2017diffusion,mori2019bounded}. To better leverage the distance information, IGNNK trains a GNN with the goal of reconstructing information on random subgraph structures (see Figure~\ref{Fig:1}). We first randomly select a subset of nodes from all available sensors and create a corresponding subgraph. We mask some nodes as missing and train the GNN to reconstruct the full signals of all nodes (including both the observed and the masked nodes) on the subgraph (Figure~\ref{fig:1first}). This training scheme allows the GNN to effectively learn the message passing mechanism, which can be further generalized to unseen nodes/graphs. Next, given observed signals from the same or even an entirely new network structure, the trained model can perform kriging through reconstruction (Figure~\ref{fig:1second}).

We compare IGNNK with other state-of-the-art kriging methods on five real-world spatiotemporal datasets. IGNNK achieves the best performance for almost all datasets, suggesting that the model can effectively generalize spatiotemporal dynamics on a sensor network. To demonstrate the transferability of IGNNK, we apply the trained models from two traffic speed datasets (METR-LA and SeData) to a new one (PeMS-Bay), and we find that the two models offer very competitive performance even when the new dataset is never seen.

\section{Related Work}

The network spatiotemporal kriging problem can be considered a special matrix completion problem in which several rows are completely missing. A common solution is to leverage the network structure as side information \cite{zhou2012kernelized,rao2015collaborative}. In a spatiotemporal setting, network kriging is similar to the problems presented in \cite{bahadori2014fast,deng2016latent,takeuchi2017autoregressive,applebykriging}. Low-rank tensor models are developed by capturing dependencies among variables \cite{bahadori2014fast} or incorporating spatial autoregressive dynamics \cite{takeuchi2017autoregressive}. Different from previous approaches, we try to gain inductive ability by using GNN. Our method is closely related the following research directions.

\paragraph{GNNs for spatiotemporal datasets} Graph Convolutional Networks (GCNs) are the most commonly used GNN. The generalized convolution operation on graphs is first introduced in \cite{bruna2014spectral}. \citet{defferrard2016convolutional} proposes to use Chebyshev polynomial filters on the eigenvalues to approximate the convolutional filters. \citet{kipf2017semi} further simplifies the graph convolution operation using the first-order approximation of Chebyshev polynomial. To model temporal dynamics on graphs, GCNs are combined with recurrent neural networks (RNNs) and temporal convolutional networks (TCNs). One of the early methods using GCNs to filter inputs and hidden states in RNNs is \cite{seo2018structured}. Later studies have integrated different convolution strategies, such as diffusion convolution \cite{li2017diffusion}, gated convolution \cite{yu2018spatio}, attention mechanism \cite{zhang2018gaan}, and graph WaveNet \cite{wu2019graph}. As these models are essentially designed for temporal forecasting problems, they are not suitable for the spatiotemporal kriging task. More recently, \citet{applebykriging} developed kriging convolutional networks (KCN), which construct a local subgraph using a nearest neighbor algorithm and train a GNN to reconstruct each individual node's static label. However, KCN ignores the fact that the labels of the node are temporally evolving in spatiotemporal datasets. Thus it requires retraining when new data is observed in the sensor networks.

\paragraph{Inductive GNNs} \citet{hamilton2017inductive} is the first to show that GNNs are capable of learning both \textit{transductive} and \textit{inductive} node representations. Some recent studies have developed inductive GNNs for recommender systems by extracting user/item embeddings on the user-item bipartite graphs. For example, \citet{zhang2019star} masks a part of the observed user and item embeddings and trains a GCN to reconstruct these masked embedding. \citet{zhang2019inductive} uses local graph patterns around a rating (i.e., an observed entry in the matrix) and builds a one-hop subgraph to train a GCN, providing inductive power to generalize to unseen users/items and transferability. \citet{Zeng2020GraphSAINT} proposes a graph sampling approach to construct subgraphs to train GNNs for large graphs. The full GNN trained using sampled subgraphs shows superior performance. Those approaches focus on applications with binary graph structure. The effects of distance in spatiotemporal data are not fully considered in these methods.

\section{Methodology}

\subsection{Problem description}
\label{PS}

Spatiotemporal kriging refers to the task of interpolating time series signals at unsampled sensor locations given signals from sampled sensor locations. We use the terms ``node'', ``sensor'', and ``location'' interchangeably throughout the paper. This study focuses on spatiotemporal kriging on networks: the spatial domain becomes an irregular network structure instead of a 2D surface. Consider a set of traffic sensors on a highway network as an example: we can model the sensors as nodes in a network, and the edges can be defined based on the typology of the highway network \cite{li2017diffusion}.

In this case, the objective of spatiotemporal kriging is to recover traffic state time series at locations with no sensors. Thus, kriging can be considered the process of setting up virtual sensors. We illustrate the real-time network kriging problem in Figure~\ref{fig:1second}. Let $\left[t_1,t_2\right]=\left\{t_1,t_1+1,\ldots,t_2-1,t_2\right\}$ denote a set of time points. Suppose we have data from $n$ sensors during a historical period $\left[1,p\right]$ ($n=8$ in Figure~\ref{fig:1second}, corresponding to sensors $\{1,\ldots,8\}$). We denote by a multivariate time series matrix $X \in \mathbb{R}^{n \times p}$ the
available data, with each row being the signal collected from a sensor. We use $X$ as training data. Let $\left[t,t+h\right)=\left\{t,t+1,\ldots,t+h-1\right\}$ be the period for which we will perform kriging. During this period, assume we have data $Y_t^s\in\mathbb{R}^{n^s_t \times h}$ available from $n^s_t$ sensors ($n^s_t=8$ in  Figure~\ref{fig:1second}, corresponding to sensors $\{2,\ldots,9\}$)), and we are interested in interpolating the signals $Y_t^u\in\mathbb{R}^{n^u_t \times h}$ on $n^u_t$ virtual sensors (i.e., new/unsampled nodes, corresponding to sensors $\{10,11\}$ in Figure~\ref{fig:1second}). Note that both $X$ and $Y_t^s$ might be corrupted with missing values. Moreover, it is possible that $n\neq n^s_t$ as some sensors may retire and new sensors may be introduced. Our goal is to estimate $Y_t^u$ given $Y_t^s$. To achieve this, we design IGNNK to learn and generalize the message passing mechanism in training data $X$.

\subsection{Subgraph signals and random masks}
\label{tra_sch}

As a first step of IGNNK, we use a random sampling procedure---Algorithm \ref{alg:A1}---to generate a set of subgraphs for training. The key idea is to randomly sample a subset of nodes to get $X_{\text{sample}}$ and build the corresponding adjacency matrix $W_{\text{sample}}$. The kriging problem is very different from the applications in recommender systems, where the graph encodes only topological information (i.e., social network or co-purchasing network). In the spatial setting, our graph is essentially fully connected, in which we expect the edge weight to diminish with Euclidean/travel distance between a pair of nodes. To better characterize the effect of ``distance'', we simply choose a purely random sampling scheme to generate sample subgraphs, instead of creating a local subgraph for each node \cite{hamilton2017inductive,applebykriging, Zeng2020GraphSAINT}. We create a mask matrix $M_{\text{sample}}$ to keep some nodes as observed and the test as ``unsampled''. We then use the generated masks $M_{\text{sample}}$, graph signals $X_{\text{sample}}$ and adjacency matrix $W_{\text{sample}}$ to train a GNN. As we can not exactly know the spatial locations of test nodes, $W_{\text{sample}}$ is also randomly generated to make the learning samples generalizable to more cases. The input data $X_{\text{sample}}$ itself could contain missing values, which we also mark as unknown. This will enable IGNNK to perform spatial interpolation under missing data scenarios.

\begin{algorithm}[!ht]
\caption{Subgraph signal and random mask generation}
\label{alg:A1}
\begin{algorithmic}[1]
\REQUIRE Historical data $X$ from sampled locations over period $[1,p]$ (size $n\times p$). \\ Parameters: window length $h$, sample size each iteration $S$, and maximum iteration $I_{\max}$. \\
\FOR {$\text{iteration} = 1:I_{\max}$}
\FOR {$\text{sample} = 1:S$}

\STATE Generate random integers $n_o$ (number of nodes selected as observed) and $n_m$ (number of nodes selected as missing) with $n_o + n_m \leq n$.

\STATE Randomly sample $n_o + n_m$ indices without replacement from $\left[1,n\right]$ to obtain $I_{\text{sample}} = \{i^1, \ldots, i^{n_o}, \ldots i^{n_o+n_m} \}$.

\STATE Randomly choose a time point $j$ within range $\left[1, p-h\right]$. Let $J_{\text{sample}} = \left[j,j+h\right)$.

\STATE Obtain submatrix signal $X_{\text{sample}} = X[I_{\text{sample}}, J_{\text{sample}}]$ with size of $\left(n_o+n_m\right)\times h$.

\STATE Construct adjacency matrix $W_{\text{sample}} \in \mathbb{R}^{(n_o + n_m) \times (n_o+n_m)}$ for nodes in $I_{\text{sample}}$.

\STATE Generate a mask matrix $M_{\text{sample}}$ of size $(n_o + n_m) \times h$, $M_{\text{sample}}[i, :] =  \begin{cases}
    1 ,     & \text{if } i \in \left[1,n_o\right],\\
    0,  &  \text{otherwise}.
  \end{cases} $
\ENDFOR
\STATE Use sets $\{X_{1:S}\}$, $\{M_{{1:S}}\}$, $\{W_{{1:S}}\}$ to train GNNs.
\ENDFOR
\end{algorithmic}
\end{algorithm}

\subsection{GNN architecture}

The second step of IGNNK is to train a GNN model to reconstruct the full matrix $X_{\text{sample}}$ on the subgraph given the incomplete signals $X^M_{\text{sample}} = X_{\text{sample}} \odot M_{\text{sample}}$, where $\odot$ denotes the Hadamard product. In order to achieve higher forecasting power, previous studies mainly combine GNNs with sequence-learning models---such as RNNs and TCNs---to jointly capture spatiotemporal correlations, in particular long-term temporal dependencies. However, for our real-time kriging task, the recovery window $h$ is relatively short. Therefore, we simply assume that all time points in the recovery window $h$ are correlated with each other, and model a length-$h$ signal as $h$ features.


Real-world spatial networks are often directed with an asymmetric distance matrix \cite[see e.g.,][]{mori2019bounded}. To characterize the stochastic nature of spatial and directional dependencies, we adopt Diffusion Graph Convolutional Networks (DGCNs) \cite{li2017diffusion} as the basic building block of our architecture:
\begin{equation}
  H_{l+1} = \sum^K_{k = 1}  T_k(\bar{W}_f) H_{l} \Theta^k_{b,l} + T_k(\bar{W}_b) H_{l} \Theta^k_{f,l},
  \label{eq:dgcn}
\end{equation}
where $\bar{W}_f = W_{\text{sample}} / \text{rowsum}(W_{\text{sample}})$ and $\bar{W}_b = W^T_{\text{sample}} / \text{rowsum}(W^T_{\text{sample}})$ are the forward transition matrix and the backward transition matrix, respectively; Here we use two transition matrices because the adjacency matrix can be asymmetrical in a directional graph. In an undirected graph, $\bar{W}_f = \bar{W}_b$. $K$ is the order of diffusion convolution; the Chebyshev polynomial is used to approximate the convolution process in DGCN, and we have $T_k (X) = 2XT_{k-1}(X)-T_{k-2}(X)$ defined in a recursive manner with $T_0(X)= I$ and $T_1(X) = X$; $\Theta^k_{f,l}$ and $\Theta^k_{b,l}$ are learning parameters of the $l$th layer that control how each node transforms received information; $H_{l+1}$ is the outputs of the $l$th layer. Unlike traditional GNNs using a fixed spatial structure, in IGNNK each sample has its own subgraph structure. Thus, the adjacency matrices $W_{\text{sample}}$ capturing neighborhood information and message passing direction are also different in different samples.

\begin{figure}[!ht]
\centering
\includegraphics[width=\columnwidth]{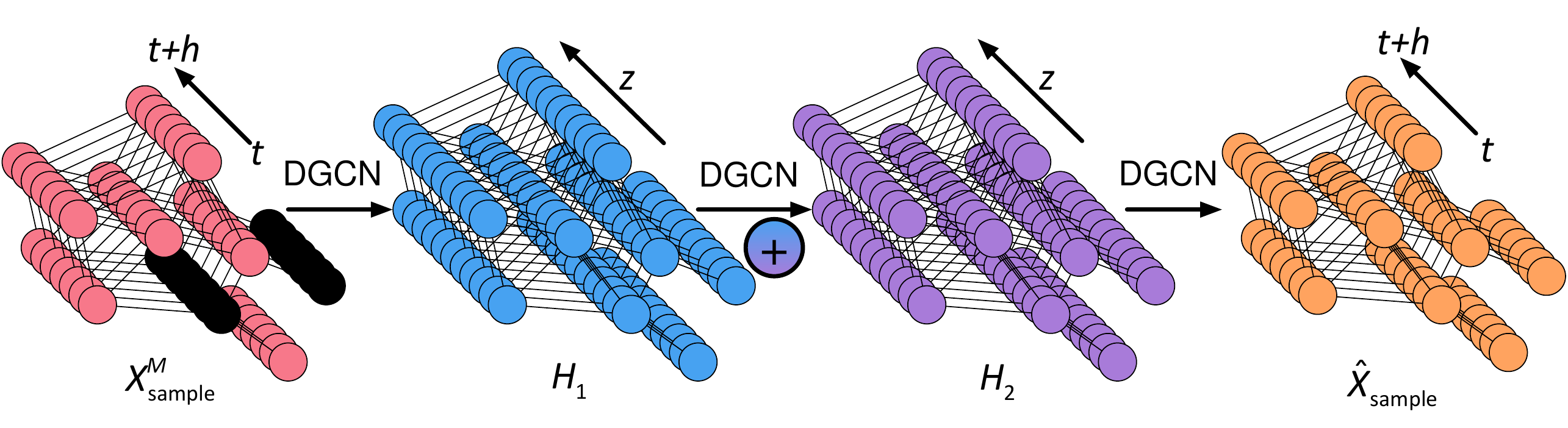}
\caption{Graph neural network structure of IGNNK}
\label{Fig:2}
\end{figure}

Figure~\ref{Fig:2} illustrates the whole graph neural networks structure of IGNNK, which is a simple 3-layer DGCN. The input to the first layer is the masked signals $H_0=X^M_{\text{sample}}$. Next, we set $H_1$ following Eq.~\eqref{eq:dgcn} with parameters $\Theta^k_{b,0} \in \mathbb{R}^{h \times z}$ and $\Theta^k_{f,0} \in \mathbb{R}^{h \times z}$. Since the masked nodes only pass $0$ to their neighbors in the first layer, a one-layer GCN cannot produce desirable features. Therefore, we add another layer of DGCN (i.e., $H_{2}$) to produce more generalized representations:
\begin{equation}
  H_{2} = \sigma\left(\sum^K_{k = 1}  T_k(\bar{W}_f) H_{1} \Theta^k_{b,1} + T_k(\bar{W}_b) H_{1} \Theta^k_{f,1}\right) + H_{1},
\end{equation}
where $\Theta^k_{f,1} \in \mathbb{R}^{z \times z}$ and $\Theta^k_{b,1} \in \mathbb{R}^{z \times z}$ are parameters of the second layer DGCN, and $\sigma(\cdot)$ is a nonlinear activation function. The reason we add $H_1$ to $H_2$ is that $H_1$ contains the information about sensors with missing data.


After obtaining the final graph representation, we use another DGCN to output the reconstruction:
\begin{equation}
 \hat{X} = \sum^K_{k = 1}  T_k(\bar{W}_f) H_{2} \Theta^k_{b,2} + T_k(\bar{W}_b) H_{2} \Theta^k_{f,2},
\end{equation}
where $\Theta^k_{f,2} \in \mathbb{R}^{z \times h}$ and $\Theta^k_{b,2} \in \mathbb{R}^{z \times h}$ are the learning parameters of the last layer. Note that we keep the structure of IGNNK as parameter-economic as possible. One can always introduce more complex structures to further improve model capacity and performance.

\subsection{Loss function and prediction}

As stated in Section \ref{PS}, the problem we address here is to recover the information of unsampled sensors. A straightforward approach is to define training loss functions only on the masked signals. However, from a methodological perspective, we prefer IGNNK be generalized to both dynamic graph structures and the unseen nodes \cite{hamilton2017inductive}. To make the learned message passing mechanism more generalized for all nodes, we use the total reconstruction error on both observed and unseen nodes as our loss function:
\begin{equation}
    J = \sum_{\text{sample}} \|\hat{X}_{\text{sample}} -  X_{\text{sample}} \|^2_F.
    \label{loss_fun}
\end{equation}

Next, we illustrate how to perform kriging to get matrix $Y_t^u$ for the virtual sensors in Figure~\ref{fig:1second}. First, we can create the $\left(n_t^s+n_t^u\right)\times h$ binary mask matrix $M_t$ and the $\left(n_t^s+n_t^u\right)\times \left(n_t^s+n_t^u\right)$ adjacency matrix $W_t$ given the typology of the sensor network during $\left[t,t+h\right)$. Then, we can feed $W_t$ and the ``masked'' signals $Y_t^M=\left[ Y_t^s; Y_t^u \right]$ with $Y_t^u=\boldsymbol{0}$ to the trained IGNNK model to get $\hat{Y}_t^M=\left[ \hat{Y}_t^s; \hat{Y}_t^u\right]$. The estimation $\hat{Y}_t^u\in \mathbb{R}^{n^u_t\times h}$ is the final kriging results for those virtual sensors.

\section{Numerical Experiments}

In this section, we assess the performance of IGNNK using five real-world spatiotemporal datasets from diverse applications: (1) \textbf{METR-LA} is a traffic speed dataset from 207 sensors in Los Angeles over four months (Mar 1, 2012 to Jun 30, 2012); (2) \textbf{NREL} registers solar power output by 137 photovoltaic power plants in Alabama state in 2006; (3) \textbf{USHCN} contains monthly precipitation of 1218 locations from 1899 to 2019; (4) \textbf{SeData} is also a traffic speed dataset collected from 323 loop detectors in the Seattle highway network; (5) \textbf{PeMS-Bay} is traffic speed dataset similar to METR-LA, which consists of a traffic speed time series of 325 sensors in the Bay Area from Jan 1, 2017 to May 13, 2017. We use PeMS-Bay to demonstrate the transferability of IGNNK. The frequency for METR-LA, NERL, SeData and PeMS-Bay are all 5-min. In terms of adjacency structure, we compute the pairwise Euclidean distance matrices for NREL and USHCN; both METR-LA and PeMS-Bay have a travel distance-based adjacency matrix (i.e., ``reachability'') which are asymmetric. For the distance-based adjacent matrix, we build the adjacency matrix following \cite{li2017diffusion}:
\begin{equation}
W_{ij} = \exp\left(-\left(\frac{\text{dist}\left(v_i , v_j\right)}{\sigma}\right)^2\right),
\label{distance_rule}
\end{equation}
where $W_{ij}$ is the adjacency matrix weights between sensors $v_i$ and $v_j$, $\text{dist}\left(v_i , v_j\right)$ represents the distance and $\sigma$ is a factor to normalize the distance. SeData has a simple binary adjacency matrix, which is built by using classical undirected graph definition:
\begin{equation}
W_{ij} =
\begin{cases}
1, & \text{if $i$ and $j$ are neighbors}, \\
0, & \text{otherwise}.
\end{cases}
\label{binary_rule}
\end{equation}

\subsection{Experiment setup}
We compare the performance of IGNNK with the following widely used kriging/interpolation and matrix/tensor factorization models. (1) \textbf{kNN}: K-nearest neighbors, which estimates the information of unknown nodes by averaging the values of the $K$ spatially closest sensors in the network. (2) \textbf{OKriging}: ordinary kriging--a well-developed spatial interpolation model \cite{cressie2015statistics}. (3) \textbf{KPMF}: Kernelized Probabilistic Matrix Factorization, which incorporates graph kernel information \cite{zhou2012kernelized} into matrix factorization using a regularized Laplacian kernel. (4) \textbf{GLTL}: Greedy Low-rank Tensor Learning, which is a transductive tensor factorization model for spatiotemporal cokriging \cite{bahadori2014fast}. GLTL deals with multiple variables using a [\textit{location}$\times$\textit{time}$\times$\textit{variable}] tensor structure. We implement a reduced matrix version of GLTL, as we only have one variable for all the datasets.

To evaluate the performance of the different models, we randomly select $\approx$25\% nodes as ``unsampled'' locations/virtual sensors ($n_t^s$) and keep the rest as observed for all five datasets. For IGNNK, we take data from the first 70\% of the time points as a training set $X$ and test the kriging performance on the following 30\% of the time points. For simplification, we keep the values of $n_m$ and $n_0$ fixed in our implementation, instead of generating random numbers as in Algorithm~\ref{alg:A1}. We choose $h=24$ (i.e., 2 h) for the three traffic datasets, $h=16$ (i.e., 80 min) for NREL, and $h=6$ (i.e., 6 months) for USHCN. We perform kriging using a sliding-window approach on: $\left[t, t+h\right)$, $\left[t+h, t+2h\right)$, $\left[t+2h, t+3h\right)$, etc. It should be noted that, since the training and test nodes are both given in the datasets, we can pre-compute the pairwise adjacency matrix for all nodes and obtain $W_{\text{sample}}$ by directly selecting a submatrix from the full adjacency matrix. 
For matrix/tensor-based models (i.e., KPMF and GLTL), we use the entire dataset as input but mask the same 25\% ``unsampled'' nodes as in IGNNK. We use the first 70\% of the time points from unsampled nodes as a validation set to select the optimal hyperparameters, and evaluate recovery performance on the following 30\% of the time points. For these models, the recovery for all 30\% of the test nodes can be generated at once, and they use more information compared to IGNNK. The full Gaussian process model for spatiotemporal kriging is very computationally expensive. Thus, we implement OKriging and kNN for each time step $t$ separately (i.e., $h=1$ without considering temporal dependencies). Again, we use the first 70\% of the time points from the unsampled sensors for validation and the remaining 30\% for testing. All the baseline models are tuned
using validataion, 
Since OKriging uses the longitude/latitude as input, it is not appropriate for road-distance-based transportation networks. Therefore, OKriging is only applied to NERL and USHCN datasets. As the adjacency matrix of SeData is given as a binary matrix instead of using actual distances, we cannot get effective neighbors for kNN. Thus, kNN is not applied to SeData.

\subsection{Experiment results}

\paragraph{Kriging performance}
Table~\ref{tab:comparison} shows the kriging performance of IGNNK and other baseline models on four datasets. As can be seen, the proposed method consistently outperforms other baseline models, providing the lowest RMSE and MAE for almost all datasets.

\begin{table*}[!ht]
\caption{Kriging performance comparison of different models on four datasets.}
\label{tab:comparison}
\centering
\begin{tabular}{crrrrrrrrrrrr}
\toprule
\multirow{2}{*}{} & \multicolumn{3}{c}{METR-LA} & \multicolumn{3}{c}{NREL} & \multicolumn{3}{c}{USHCN} & \multicolumn{3}{c}{SeData} \\
\cmidrule{2-13}
Model & RMSE & MAE & $R^2$ & RMSE & MAE & $R^2$  & RMSE & MAE & $R^2$ & RMSE & MAE & $R^2$ \\
\midrule
IGNNK & \textbf{9.048} & \textbf{5.941} & \textbf{0.827} & \textbf{3.261} & \textbf{1.597} & \textbf{0.885} & \textbf{3.205} & 2.063 & \textbf{0.771} & \textbf{6.863} & \textbf{4.241} & \textbf{0.537} \\
kNN & 11.071 & 6.927 & 0.741 & 4.192 & 2.850 &0.810& 3.400 & 2.086 & 0.742 & - & - & -\\
KPMF & 12.851 & 7.890 &0.652 & 8.771 & 7.408 &0.169 & 6.663  & 4.847 &0.011 & 13.060 & 8.339&-0.673 \\
GLTL & 9.668 &  6.559&0.803 & 4.840 &  3.372 &0.747 & 5.047 & 3.396&0.432 & 6.989 & 4.285&0.520 \\
OKriging & - & - & - & 3.470 & 2.381 &0.869 & 3.231 & \textbf{1.999} &0.767& -& - & -\\
\bottomrule
\end{tabular}
\end{table*}

As can be seen, IGNNK achieves good performance on the two spatial datasets---NREL and USHCN where OKriging fits the tasks very well. There are two cases worth mentioning in Table~\ref{tab:comparison}. First, kNN and OKriging also give comparable results to IGNNK on USHCN data, the MAE of OKriging is even lower than the one of IGNNK. This is due to the fact that sensors have a high density and precipitation often shows smooth spatial variation.
In this case, local spatial consistency might be powerful enough for kriging, and thus we do not see much improvement from IGNNK. For SeData, GLTL also shows good performance. A potential reason is the data limitation: the adjacency structure of SeData is given as a binary matrix of sensor connectivity (i.e., 1 if two sensors are neighbors next to each other on a highway segment, and 0 otherwise;). In this case, $W$ only encodes the typology, instead of the full pairwise distance information as in the other datasets. In fact, relative distance serves a critical role in the kriging task on a highway network due to the complex causal structures and dynamics of traffic flow \cite{li2017diffusion}. As a result, we do expect IGNNK to be less powerful due to the lack of ``distance'' effect, and all the methods give lower $R^2$ on SeData.

\begin{figure}[!ht]
\centering
\includegraphics[width=1.05\columnwidth]{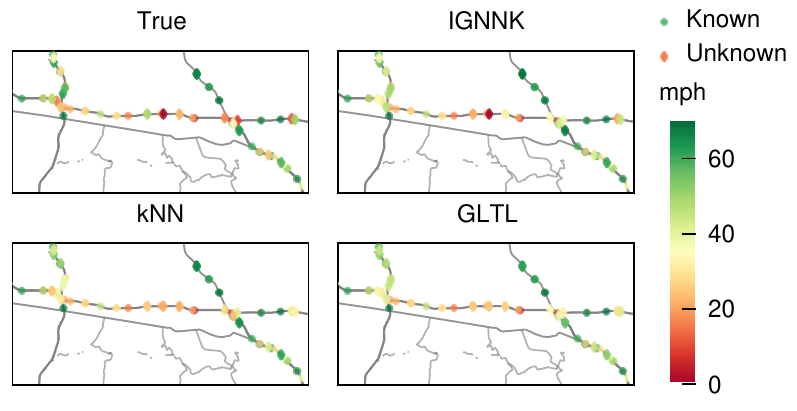}
\caption{Spatial presentation of kriging performance in a crowded evening-peak time point (2012-05-22 17:45) of the METR-LA dataset. The speed values of ground truth, IGNNK, kNN, and GLTL are visualized on the top-left, top-right, bottom-left, and bottom-right, respectively.}
\label{Fig:ck_metr}
\end{figure}

We provide an example of spatial visualizations of IGNNK and other baseline models in Figure. \ref{Fig:ck_metr}. It can be seen that the reconstruction (star) of IGNNK is closer to ground truth compared to kNN and GLTL. 

\paragraph{Transfer learning performance}

Next, we demonstrate the transferability of IGNNK. Our experiments are based on three datasets---METR-LA, SeData, and PeMS-Bay---collected from three different highway networks using the same type of sensors (i.e., loop detectors).

The PeMS-Bay dataset is very similar to METR-LA: both datasets register traffic speed with a 5-min frequency, and both of them provide travel distance for each of the sensors. The two datasets are used side by side in \cite{li2017diffusion}. SeData has the same format; however, as mentioned, the dataset provides a simple binary adjacency matrix showing connectivity. To show the effect of the adjacency matrix, we train two sets of models separately following the same approach as the kriging experiments: one with a Gaussian kernel adjacency matrix based on pairwise travel distance (same as METR-LA), and the other on the binary connectivity matrix (same as SeData).

\begin{table*}[!ht]
\caption{Kriging performance of different models on PeMS-Bay. The last two rows shows the transferability of the two IGNNK models trained on METR-LA and SeData. }
\label{tab:transfer}
\centering
\begin{tabular}{crrrrrrrr}
\toprule
\multirow{2}{*}{}  & \multicolumn{4}{c}{Gaussian} & \multicolumn{4}{c}{Binary}\\
\cmidrule(lr){2-5} \cmidrule(lr){6-9}
Model & RMSE & MAE  & MAPE & $R^2$ & RMSE  & MAE & MAPE & $R^2$ \\

\midrule
IGNNK & \textbf{6.093} & \textbf{3.663}  & \textbf{8.16}\% &\textbf{0.574} & 9.245 & 5.394 & 13.26\% & 0.161\\
kNN & 7.431 & 4.245 & 9.13\% &0.458 & - & - & - & - \\
KPMF & 7.332 & 4.293 & 9.21\% &0.472 & 10.065  & 5.985 & 16.03\% & 0.005\\
GLTL & 8.846 & 4.486 & 10.25\% &0.232 & 8.504 & 4.962 & 12.24\% & 0.290 \\
\midrule
\multirow{2}{*}{IGNNK Transfer} & \multicolumn{4}{c}{METR-LA} & \multicolumn{4}{c}{SeData}\\
\cmidrule(lr){2-5} \cmidrule(lr){6-9}
& \textbf{6.713} & \textbf{4.173}  & \textbf{9.19}\%&\textbf{0.525} & 11.484 & 6.456  & 15.10\%&-0.388\\
\bottomrule
\end{tabular}
\end{table*}

\begin{figure}[!ht]
\centering
\includegraphics{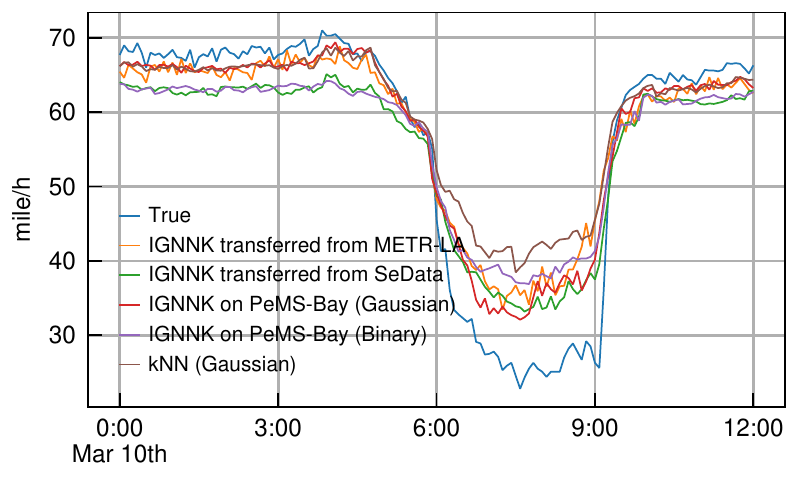} 
\caption{Kriging performance on unknown nodes in PeMS-Bay dataset.} 
\label{Fig:transfer_learning}
\end{figure}

The top part of Table~\ref{tab:transfer} shows the results obtained using Gaussian kernel adjacency (\textit{Gaussian}) and binary adjacency (\textit{Binary}), respectively. Not surprisingly, IGNNK-Gaussian offers the best accuracy, clearly outperforming all other models.  In general, \textit{Gaussian} offers better performance than \textit{Binary} for all the baseline models. Similar to the kriging experiment on SeData, we find that \textit{Binary} IGNNK and \textit{Binary} GLTL demonstrate comparable performance.

We then apply the two IGNNK models---trained on METR-LA and SeData, respectively---directly on the test data of PeMS-Bay and obtain the kriging errors (the last two rows in Table~\ref{tab:transfer}). As can be seen, IGNNK (METR-LA) outperforms other baselines except the IGNNK trained on PeMS-Bay; however, IGNNK (SeData) does not offer encouraging results in this case, its $R^2$ is below 0, meaning that the transfer model is even worse than the mean predictor. Our experiment shows that IGNNK can effectively learn the effect of pairwise distance from METR-LA and then generalize the results to PeMS-Bay. Figure~\ref{Fig:transfer_learning} shows an example of kriging results of these models. While no models can reproduce the sudden drop during 6:00 am--9:00 am, we can see that \textit{Gaussian} provides much better recovery during non-peak hours than \textit{Binary}. The analysis on transferability further confirms the critical role of distance in kriging tasks.

\paragraph{Virtual sensors}

\begin{figure*}[!ht]
\centering
\subfigure[Virtual sensors/Kriging for METR-LA between the 1st node and the 177th node]{
\label{Fig:metr_virtual}
\includegraphics{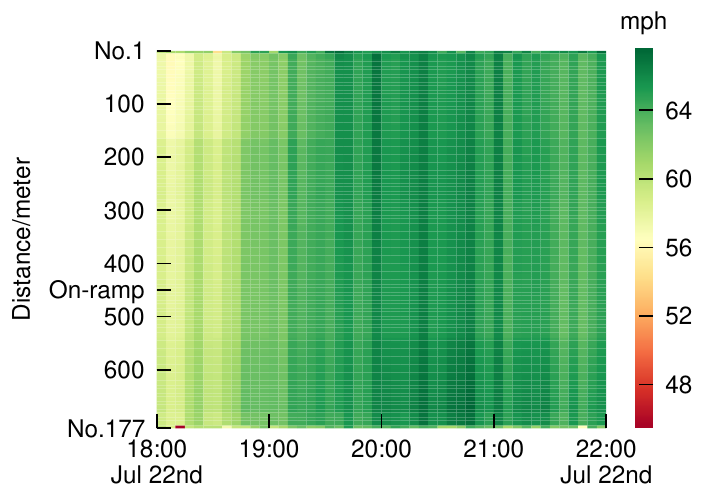}} \hspace{0.5cm}
\subfigure[Virtual sensors/Kriging for USHCN data between the 8th sensor and the 10th sensor]{
\label{Fig:u_virtual}
\includegraphics{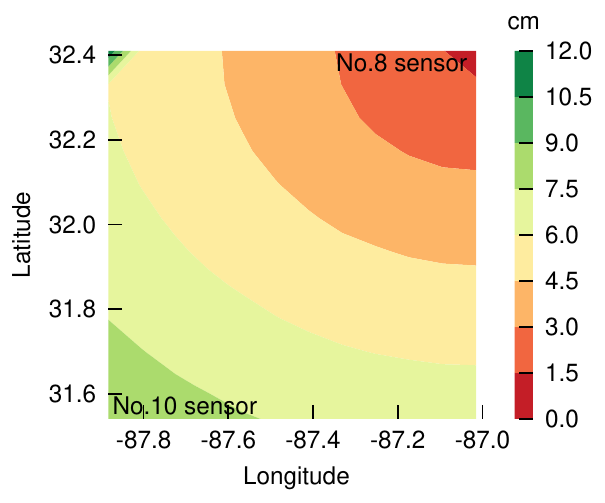}}
\caption{Visualization of Kriging results on virtual sensors.}
\label{Fig:virtual_sensors}
\end{figure*}

Given a location and its surrounding sensors' data, IGNNK can be used to generate virtual sensors for the target location. Figure~\ref{Fig:virtual_sensors} shows the outputs of virtual sensors on METR-LA and USHCN datasets. The outputs are produced by IGNNK models trained on METR-LA and USHCN. For METR-LA, we place non-existent sensors every 5 meters between two adjacent sensors (No 1 and No 177) on the road. From Figure~\ref{Fig:metr_virtual}, we can find that the values of virtual sensors produced by IGNNK smoothly vary from the value of the 1st sensor to the one of the 177th sensor. At 18:10 Jul 22nd, the speed of 177th sensor dramatically drops below 48 mph. The drops might be caused by the
heavy on-flows at the on-ramp between those two sensors.
In this case, virtual sensors produced by IGNNK fail to produce a smooth transition, as they still output relatively high speed. This is because the dynamic graph of METR-LA is constructed according to the road distance. Other factors like on-ramp/off-ramp information are not incorporated into this model. In Figure~\ref{Fig:u_virtual}, we place 100 virtual sensors between the 8th location and the 10th location. At the 8th location, the precipitation is 0cm. While at the 10th location, the precipitation is above 9 cm. We can find that the precipitation at virtual sensors near the 8th location are lower, and the precipitation at virtual sensors near 10th location are higher. Notably, the values of virtual sensors are also affected by other sensors at long distances. Generally, we find that IGNNK learns to generate values according to the distance information and real values of its surrounding sensors.

\section{Conclusion}

In this paper, we introduce IGNNK as a novel framework for spatiotemporal kriging. Instead of learning transductive latent features, the training scheme provides IGNNK with additional generalization and inductive power. Thus, we can apply a trained model directly to perform kriging for any new locations of interest without retraining. Our numerical experiments show that IGNNK consistently outperforms other baseline models on five real-world spatiotemporal datasets. Besides, IGNNK demonstrates remarkable transferability in our examlple of a traffic data kriging task. Our results also suggest that ``distance'' information in graphs plays a critical role in spatiotemporal kriging, which is different from applications in recommender systems where a graph essentially encodes topological information. The flexibility of this model allows us to model time-varying systems, such as moving sensors (e.g., probe vehicles) or crowdsourcing systems, which will create a dynamic network structure.

There are several directions for future work. First, we can adapt IGNNK to accommodate multivariate datasets as a spatiotemporal tensor \cite[e.g.,][]{bahadori2014fast}. Second, better temporal models, such as RNNs and TCNs, can be incorporated to characterize complex temporal dependencies. This will allows us to perform kriging for a much longer time window with better temporal dynamics. Third, one can further combine time series forecasting with kriging in an integrated framework, providing forecasting results for both existing and virtual sensors for better decision making.

\section*{Acknowledgments}
This research is supported by the Natural Sciences and Engineering Research Council (NSERC) of Canada and the Canada Foundation for Innovation (CFI). Y. Wu would like to thank the Institute for Data Valorization (IVADO) for providing a scholarship to support this study. The authors declare no competing financial interests with this work.

\bibliographystyle{aaai21}
\bibliography{refs}
\end{document}